\documentclass[conference]{IEEEtran}
\IEEEoverridecommandlockouts

\usepackage{cite}
\usepackage{amsmath,amssymb,amsfonts}
\usepackage{algorithmic}
\usepackage{graphicx}
\usepackage{textcomp}
\usepackage{booktabs}
\usepackage{xcolor}
\usepackage{hyperref}
\usepackage{multicol}
\usepackage{multirow}
\def\BibTeX{{\rm B\kern-.05em{\sc i\kern-.025em b}\kern-.08em
    T\kern-.1667em\lower.7ex\hbox{E}\kern-.125emX}}
\begin{document}

\title{Memory-Free and Parallel Computation for Quantized Spiking Neural Networks\\}

\author{
    \IEEEauthorblockN{ Dehao Zhang$^{1}$, Shuai Wang$^{1}$, Yichen Xiao$^{1}$, Wenjie Wei$^{1}$, Yimeng Shan$^{1}$, Malu Zhang$^{1,*}$, Yang Yang$^{1}$\thanks{*Corresponding author}}
    \IEEEauthorblockA{$^{1}$School of Computer Science and Engineering, University of Electronic Science and Technology of China
    }
    }
    
\maketitle

\begin{abstract}
Quantized Spiking Neural Networks (QSNNs) offer superior energy efficiency and are well-suited for deployment on resource-limited edge devices. However, limited bit-width weight and membrane potential result in a notable performance decline. In this study, we first identify a new underlying cause for this decline: the loss of historical information due to the quantized membrane potential. To tackle this issue, we introduce a memory-free quantization method that captures all historical information without directly storing membrane potentials, resulting in better performance with less memory requirements. To further improve the computational efficiency, we propose a parallel training and asynchronous inference framework that greatly increases training speed and energy efficiency. We combine the proposed memory-free quantization and parallel computation methods to develop a high-performance and efficient QSNN, named MFP-QSNN. Extensive experiments show that our MFP-QSNN achieves state-of-the-art performance on various static and neuromorphic image datasets, requiring less memory and faster training speeds. The efficiency and efficacy of the MFP-QSNN highlight its potential for energy-efficient neuromorphic computing.

\end{abstract}


\section{Introduction}

Spiking Neural Networks (SNNs)~\cite{maass1997networks,gerstner2002spiking,izhikevich2003simple} employ effective neuronal dynamics and sparse spiking activities to mimic the biological information processing mechanisms closely. Within this framework, spiking neurons compute only upon the arrival of input spikes and remain silent otherwise. This event-driven mechanism~\cite{caviglia2014asynchronous} ensures sparse accumulate (AC) operations within the SNNs, significantly reducing the burden of extensive floating-point multiply-accumulate (MAC) operations~\cite{orchard2015converting}. However, with the development of deep SNN learning algorithms~\cite{stockl2021optimized, shen2024efficient,fang2021incorporating, zhu2024online, shen2023esl} and larger network architectures~\cite{shan2024advancing,zhu2024tcja,yao2023attention,yao2024spike,shen2021hybridsnn}, the complexity and memory requirements of SNNs significantly increase. This contradicts the objective of energy efficiency and application in edge computing.

To further reduce the memory requirements and energy consumption of SNNs, substantial research~\cite{hu2023fast, li2022quantization, deng2021comprehensive, chowdhury2021spatio} explore Quantized Spiking Neural Networks (QSNNs). Deng et al.~\cite{deng2021comprehensive} optimize a pre-trained full-precision SNN using the ADMM method for low-precision weight quantization. Concurrently, Chowdhury et al.~\cite{chowdhury2021spatio} apply K-means clustering quantization to maintain reasonable precision with 5-bit synaptic weights in SNNs. These methods effectively reduce the computational burden of full-precision synaptic operations. However, they ignore the critical role of optimizing the memory requirements of membrane potentials. This limitation restricts the potential for QSNNs to enhance energy efficiency and computational performance.

Subsequently, some research~\cite{yin2024mint,wei2024q,wang2024ternary}
introduce a dual quantization strategy that effectively quantizes synaptic weights and membrane potentials into low bit-width integer representation. These methods not only further reduce the memory requirements and computational consumption but also simplify the hardware logic through lower precision AC operations~\cite{liu2023low}. These improvements make QSNNs particularly well-suited for efficient deployment on resource-limited edge devices. However, as the bit-width of membrane potentials is reduced to 1/2 bits, it leads to a precipitous decline~\cite{sulaiman2020weight} in performance. Therefore, exploring efficient quantization methods for SNNs that maintain high performance at low bit-width of membrane potentials remains a critical challenge.

In this study, we thoroughly analyze the causes of performance decline in those methods. The primary issue is that limited bit-width membrane potentials in QSNNs can only retain historical information for 1/2 timesteps. To address this challenge, we propose a memory-free and parallel computation method for QSNN (MFP-QSNN). It not only preserves the efficient temporal interaction and asynchronous inference capabilities of SNNs but also significantly reduces their memory and computational resource requirements. Extensive experiments are conducted on static image
and neuromorphic datasets demonstrate that our MFP-QSNN outperforms other QSNN models, with lower memory requirements and faster training speeds. 
Our method offers a novel approach for achieving lighter-weight and higher-performance edge computing. The main contributions are summarized as follows:

\begin{itemize}
\item We find that quantizing membrane potentials into low bit-widths results in the loss of historical information, significantly reducing the spatio-temporal interaction capabilities of QSNNs. This is a major reason for the performance decline in QSNNs.
\item We propose a high-performance and efficient spiking model named MFP-QSNN. This model effectively retains historical information to enhance accuracy with reduced memory requirements and utilizes parallel processing to significantly boost computational efficiency.
\item Extensive experiments are conducted to evaluate the performance of the proposed MFP-QSNN on static and neuromorphic datasets. The results show that our method achieves state-of-the-art accuracy while requiring less memory and enabling faster training speeds.
\end{itemize}

\section{Preliminary}
\subsection{Leaky Integrate-and-Fire model}
SNNs encode information through binary spikes over time and work in an event-driven mechanism, offering significant energy efficiency. As fundamental units of SNNs, various spiking models~\cite{zhang2024tc, song2024spiking, shaban2021adaptive, zhang2024spike} are developed to mimic biological neuron mechanisms. The Leaky Integrate-and-fire (LIF) model is considered to be the most effective combination of biological interpretability and energy efficiency, defined as follows:
\begin{equation}
H[t] = \tau U[t-1] + W S[t].\label{1}
\end{equation}

Here, \(\tau\) denotes the membrane time constant, and \(S[t]\) represents the input spikes at time step \(t\). If the presynaptic membrane potential \(H[t]\) surpasses the threshold \(V_{th}\), the spiking neuron fires a spike \(S[t]\). \(U[t]\) is the membrane potential. It retains the value \(H[t]\) if no spike is generated, and reverts to the reset potential \(V_{reset}\) otherwise. The spiking and reset mechanisms are illustrated by Eq~\ref{2} and Eq.~\ref{3}:

\begin{equation}
S[t] = \Theta \left(H[t]-V_{th}\right), \label{2}
\end{equation}
\begin{equation}
U[t] = H[t] \left(1-S[t]\right)+ V_{rest}S[t], \label{3}
\end{equation}
where \( \Theta \) denotes the 
Heaviside function, defined as 1 for \(v \geq 0\) and 0 otherwise. Generally, \(H[t]\) serves as an intermediate state in computations and does not require dedicated storage. Instead, \(U[t]\) must be stored to ensure that the network retains historical membrane potential information for learning. However, as the scale of the network expands, the 32-bit full-precision \( W \) and \( U \) become significant barriers to deploying SNNs on edge devices~\cite{roy2019scaling}.

\subsection{Quantized Spiking Neural Networks}

To enhance the energy efficiency of SNNs, some research~\cite{yin2024mint,wei2024q,wang2024ternary} suggest quantizing $W$ and $U$ to lower bit-widths, thereby substantially reducing the memory and computational requirements. Among them, uniform quantization is commonly employed, defined as follows:

\begin{equation}
\mathcal{Q}(\alpha, b) = \alpha \times \left\{0, \pm\frac{1}{2^{b-1}-1}, \pm\frac{2}{2^{b-1}-1}, \ldots, \pm1\right\},\label{4} 
\end{equation}
where \( \alpha \) represents the quantization factor, usually expressed as a 32-bit full-precision value. \( b \) denotes the number of bits used for quantization. The accumulation of membrane potential in QSNNs, following the uniform quantization of $W$ and $U$, is described by Eq.~\ref{3} as follows:

\begin{equation}
\alpha_1 \hat{H}[t] = \alpha_2 \tau \hat{U}[t-1] + \alpha_3 \hat{W} S[t].\label{5} 
\end{equation}
 
After quantization, the membrane potentials $\hat{H}[t]$ and $\hat{U}[t]$, along with the synaptic weights $\hat{W}$, are represented as integer values. $\alpha_i$ denotes different full-precision quantization factors. Building on this, Yin et al.~\cite{yin2024mint} and Wang et al.~\cite{wang2024ternary} explored the relationships between various scaling factors $\alpha$. By incorporating different $\alpha$ values into $V_{th}$, they eliminated potential MAC operations during inference. These approaches effectively reduce the deployment challenges of SNNs on edge devices. However, significant performance degradation occurs when membrane potentials are quantized to 1-2 bits, indicating substantial room for improvement in lightweight SNN quantization strategies.
\begin{figure}[htpb] 
    \centering 
    \includegraphics[scale=0.42]{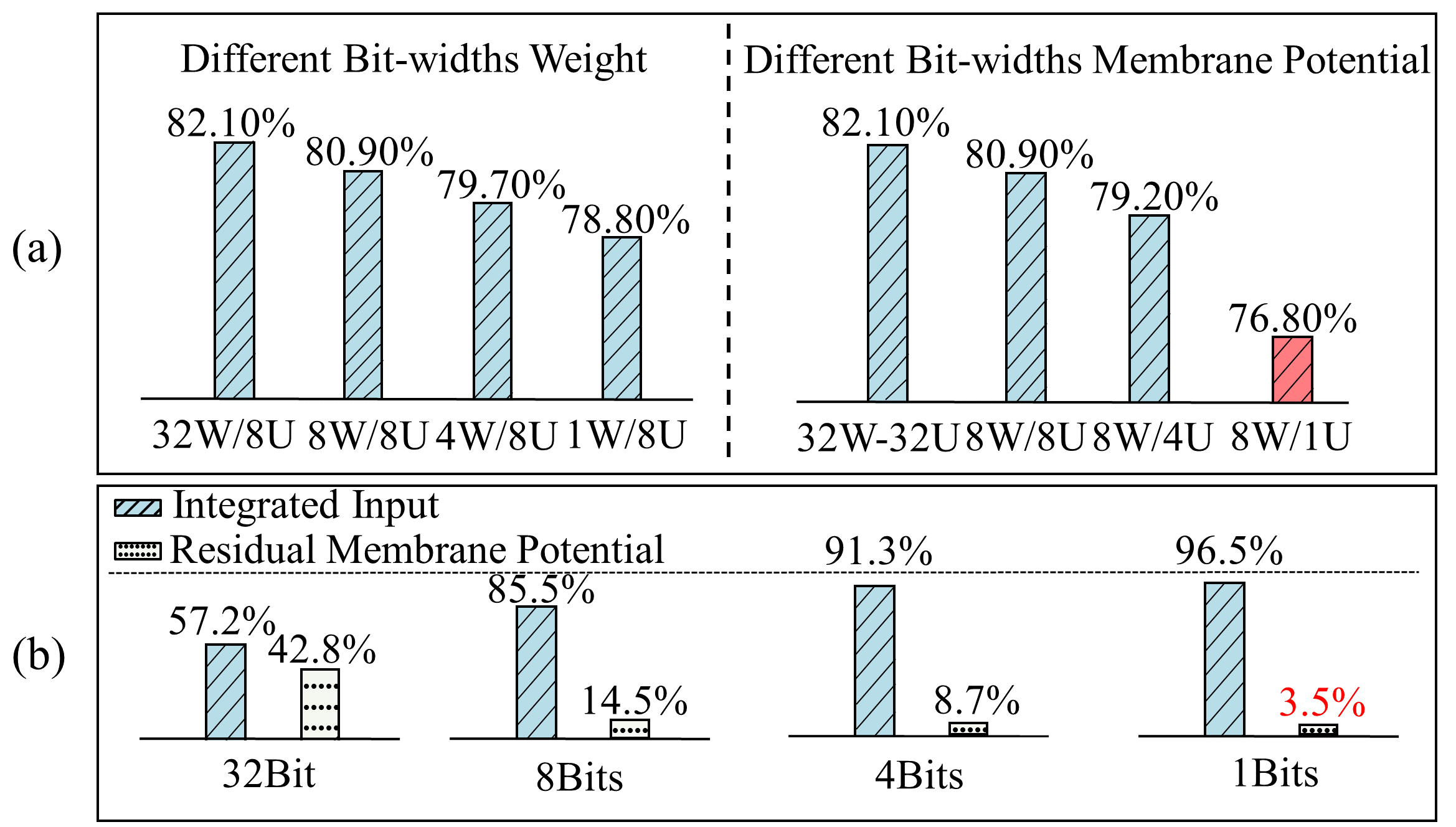}
    \centering
    \caption{Limited spatio-temporal interaction: (a) The performance under different bit-widths weights and membrane potentials. (b) The proportion of spike inputs and residual membrane voltages under different bit-widths.}
    \label{analysis}
\end{figure}
\section{Method}
In this section, we first analyze the reasons behind the significant performance degradation of QSNNs. Building on these insights, we introduce a novel memory-free and parallel computation for QSNN (MFP-QSNN), which incorporates a memory-free quantization strategy alongside a parallel training and asynchronous inference framework. MFP-QSNN achieves enhanced performance with lower memory requirements.
\subsection{Problem Analyze}
As shown in Figure \ref{analysis}.(a), we evaluate the impact of different bit widths of synaptic weights and membrane potentials on the performance of QSNN under CIFAR10DVS datasets. When the model is quantized to 8 bits, there is no significant change in accuracy, hence we select this setting as the baseline. Regarding the membrane potential, model accuracy is minimally impacted even if the weights are quantified to binary. Conversely, when synaptic weights are fixed at 8 bits and only the membrane potential is quantized to lower bit-widths, QSNNs' performance is significantly reduced. Further analysis of Eq.~\ref{5} reveals that 1/2 bits membrane potential, under the influence of the decay factor \(\tau\), may decay to zero after at most two shifts. This limitation restricts the retention of historical information to merely two timesteps, thereby impairing the neuron's ability to process spatiotemporal information.
To further validate this hypothesis, we analyzed the proportionate relationship between \( U[t-1] \) and \( W S[t] \) in \( H[t] \) across different bit-width conditions. As shown in Fig.\ref{analysis}.(b), spikes are primarily triggered by current inputs, particularly under conditions of low bits. Therefore, the reduced bit-width of the membrane potential is a critical limiting factor in QSNN.

\begin{figure*}[htpb] 
    \centering 
    \includegraphics[scale=0.5]{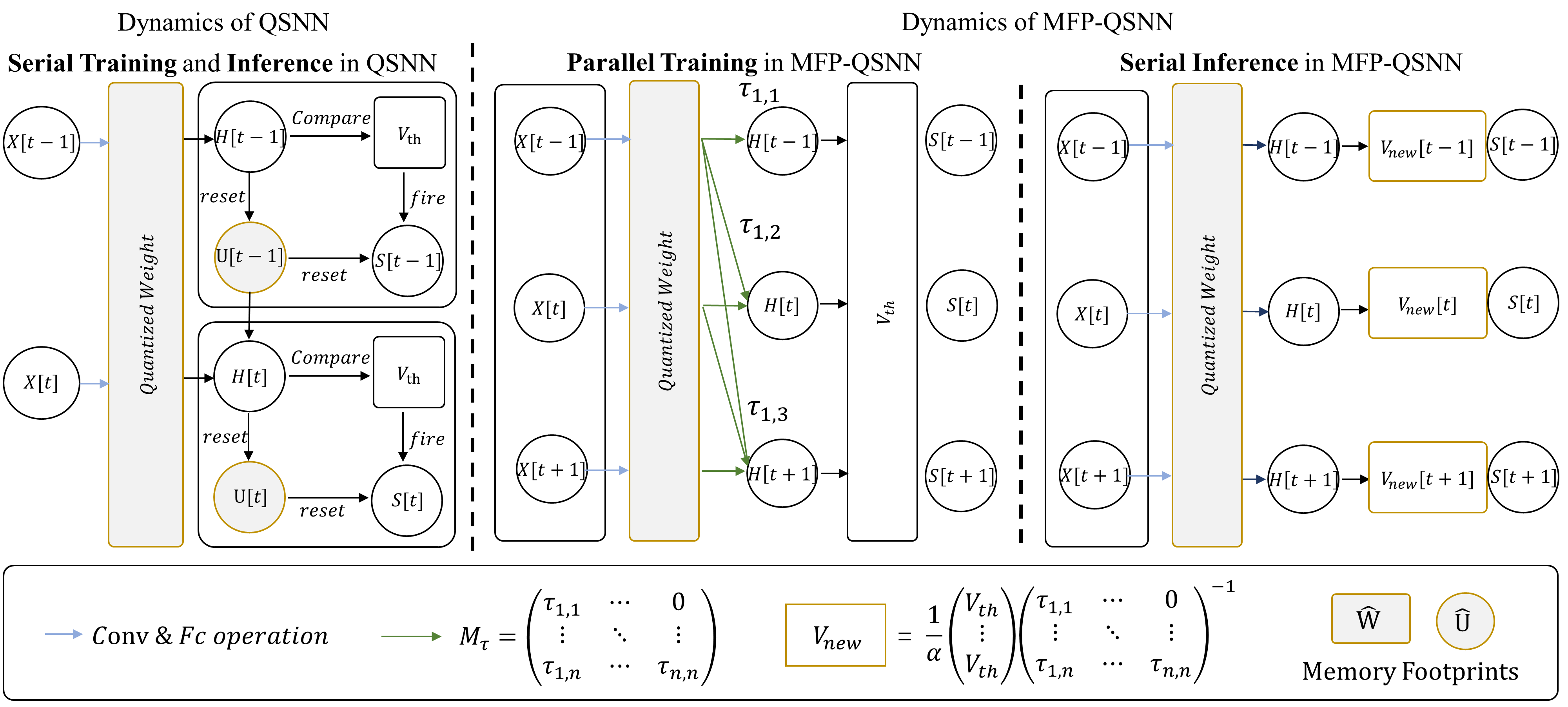}
    \centering
    \caption{Comparative analysis of dynamics between LIF neurons and our MFP-QSNN. During the training phase, $M_{\tau}$ ensures that MFP-QSNN supports parallel training without the explicit need to store U[t-1] for historical information exchange. In the inference phase, $M_{\tau}$ is integrated into the $V_{th}$ at each timestep, maintaining asynchronous inference characteristics. }
    \label{all}
\end{figure*}

\subsection{Memory-free Quantization}

To further reduce the memory requirements of QSNNs while retaining the efficient spatio-temporal interaction capabilities of SNNs, we introduce an innovative quantization method for updating membrane potentials, described as follows:
\begin{equation}
H[t] = \alpha \sum_{i=1}^{t} \tau_i \hat{W} S[i].\label{6} 
\end{equation}

Here, \(\alpha\) denotes the quantization factor of weights, and the decay factor \(\tau_i\)  is designed as a time-dependent learnable variable. Unlike Eq.\ref{1}, $H[t]$ does not explicitly rely on \(U[t-1]\) for spatio-temporal interaction but is derived by directly summing synaptic inputs from previous timesteps. Consequently, this approach obviates the need for membrane potentials, thereby reducing memory usage during inference. Additionally, \(\sum_{i=1}^{t} \tau_i \hat{W} S[i]\) ensures that \(H[t]\) captures all historical information, with \(\tau_i\) dynamically adjusting the contributions from each previous timesteps. It further significantly enhances the spatio-temporal interaction capabilities of the QSNNs.

\subsection{Parallelized Network Training and Serial Inference}

In Eq.6, \( H[t] \) requires the computation of presynaptic inputs from previous time steps [1, 2, \(\cdots\), t-1]. Consequently, commonly used serial training methods~\cite{wu2019direct, wu2018spatio} substantially increase both the network's computational customization and the training time.
Therefore, we introduce an efficient parallel network training approach, which is defined as follows:

\begin{equation}
\begin{cases}
\hat{\textbf{H}} = \alpha \textbf{M}_\tau \hat{W}\textbf{S}, & \textbf{M}_\tau \in R^{T \times T},\\ 
\textbf{S} = \Theta \left( \hat{\textbf{H}} - V_{th} \right) , & \textbf{S} \in 
\{0,1\}^T, \label{7}
\end{cases}
\end{equation}
\( \hat{\textbf{H}}, \textbf{S} \in \mathbb{R}^{T \times B \times N} \) represent the quantized presynaptic membrane potential and the spikes, respectively. Specifically, \( T \) represents the timestep, \( B \) denotes the batch size, and \( N \) is the dimension. The matrix \( \textbf{M}_\tau \) is a \( T \times T \) matrix, which is further constrained to a lower triangular form to ensure that the information at time \( t \) is only dependent on information from previous timesteps. It can be described as follows:
\begin{equation}
\textbf{M}_\tau=\left(\begin{matrix}\tau_{11}&\cdots&0\\\vdots&\ddots&\vdots\\\tau_{n1}&\cdots&\tau_{nn}\\\end{matrix}\right),
\label{8}
\end{equation}
as shown in Fig.\ref{all}, 
both \( H \) and \( S \) can be directly obtained through a single matrix operation, thereby significantly enhancing the network's training speed.

To accommodate the asynchronous inference capabilities of SNNs, MFP-QSNNs should retain the same serial computational characteristics as described in Eq.\ref{1}. Consequently, we decouple the parallel training matrix \( \textbf{M}_\tau \) and $\alpha$ into the threshold of each time step. In this manner, we can attain the same inference speed as prior SNNs. The dynamics equation can be described by Eq.~\ref{9}:

\begin{equation}
\begin{cases}
&\bar{H}^l[t] = \hat{W}^l S^{l-1}[t], \textbf{        }\hat{W}^l \in \{-1, +1\}^T, \\ 
&S^{l}[t] =
\begin{cases}
1, \textbf{        }\bar{H}^l[t] \geq V_{th} \textbf{M}_{\tau}^{-1}[i] / \alpha, \\ 
0, \textbf{        }\bar{H}^l[t] < V_{th} \textbf{M}_{\tau}^{-1}[i] / \alpha.
\end{cases}
\end{cases}\label{9}
\end{equation}

\( \textbf{M}_\tau^{-1} \) denotes the inverse of matrix \( \textbf{M}_\tau \). Given that \(\det{(\textbf{M}_\tau)} \neq 0\), the inverse of matrix \( \textbf{M}_\tau \) definitely exists. In Eq.~\ref{9}, the spiking firing process depends solely on the current values of \(H[t]\) and \(V_{th}\), eliminating the need for historical information. Thus, MFP-QSNN ensures the capability for asynchronous inference. Combined with parallel training and asynchronous inference, MFP-QSNN achieves enhanced spatio-temporal information interaction with lower computational complexity and faster training speed.

\section{Experiment}
We conduct extensive experiments on various datasets. To address the non-differentiability of spikes, we employed surrogate gradients (SG) methods~\cite{neftci2019surrogate}. Extensive experiments show that our MFP-QSNN method exhibits higher performance. Additionally, ablation studies further demonstrate that our approach significantly enhances the spatio-temporal interaction capabilities and training speed in QSNN. 

\subsection{Compare with SOTA models}
We evaluate the MFP-QSNN method across various types of image datasets, including static datasets such as CIFAR~\cite{krizhevsky2009learning}, TinyImageNet~\cite{deng2009imagenet}, and the neuromorphic dataset CIFAR10DVS~\cite{li2017cifar10}. Specifically, weights are quantized to binary form through Wei et al~\cite{wei2024q}. As shown in Table~\ref{table1}, MFP-QSNN achieves the highest Top-1 accuracy among existing similar quantization methods, further narrowing the gap with full-precision SNNs. For static datasets, our method attaines an accuracy of 95.90\% on CIFAR10. Additionally, our method achieves an accuracy of 55.6\% on TinyImageNet, with only a 0.9\% gap compared to full-precision SNNs. For neuromorphic datasets, our method reaches an accuracy of 81.1\% on CIFAR10DVS. This highlights that MFP-QSNN can enhance the spatiotemporal interaction capabilities of neurons.

\begin{table}[t]
\caption{Classification performance comparison on both static image datasets and neuromorphic datasets.}
\label{table1}
\centering
\begin{tabular}{@{}ccccc@{}}
\toprule
Methods    & Architecture & Bits (W/U) & Timesteps & Acc (\( \% \)) \\ \midrule
\multicolumn{5}{c}{Statics CIFAR10 Dataset}\\[1ex]
TET~\cite{deng2022temporal} & ResNet19 & 32/32 & 4 & 96.3 \\
TCDSNN~\cite{zhou2021temporal} & VGG16 & 2/32 & - & 90.9 \\
MINT~\cite{yin2024mint} & ResNet19 & 2/2 & 4 & 90.7 \\
Ours & ResNet19 & 1/- & 4 & 95.9 \\ \midrule 
\multicolumn{5}{c}{Statics CIFAR100 Dataset}\\[1ex]
TET~\cite{deng2022temporal} & ResNet19 & 32/32 & 4 & 79.5 \\ 
ALBSNN~\cite{pei2023albsnn} & 6Conv1FC & 1/32 &  4 & 69.5 \\
CBP-QSNN~\cite{yoo2023cbp} & VGG16 & 1/32 & 32 & 66.5 \\
Ours & ResNet19 & 1/- & 4 & 79.1 \\ \midrule 
\multicolumn{5}{c}{Statics TinyImageNet Dataset}\\[1ex]
TET~\cite{deng2022temporal} & VGG16 & 32/32 & 4 & 56.7 \\ 
MINT~\cite{yin2024mint} & VGG16 & 2/2 &  4 & 48.6 \\
Q-SNN~\cite{wei2024q} & VGG16 & 1/2 & 4 & 55.0 \\
Ours & VGG16 & 1/- & 4 & 55.6 \\ \midrule 
\multicolumn{5}{c}{Neuromorphic CIFAR10DVS Dataset}\\[1ex]
TET~\cite{deng2022temporal} & VGGSNN & 32/32 & 10 & 82.1 \\ 
Q-SNN~\cite{wei2024q} & VGGSNN & 1/2 & 10 & 80.0 \\
Ours & VGGSNN & 1/- & 10 & 81.1 \\ \bottomrule          
\end{tabular}
\end{table}

\begin{figure}[htpb] 
    \centering 
    \includegraphics[scale=0.35]{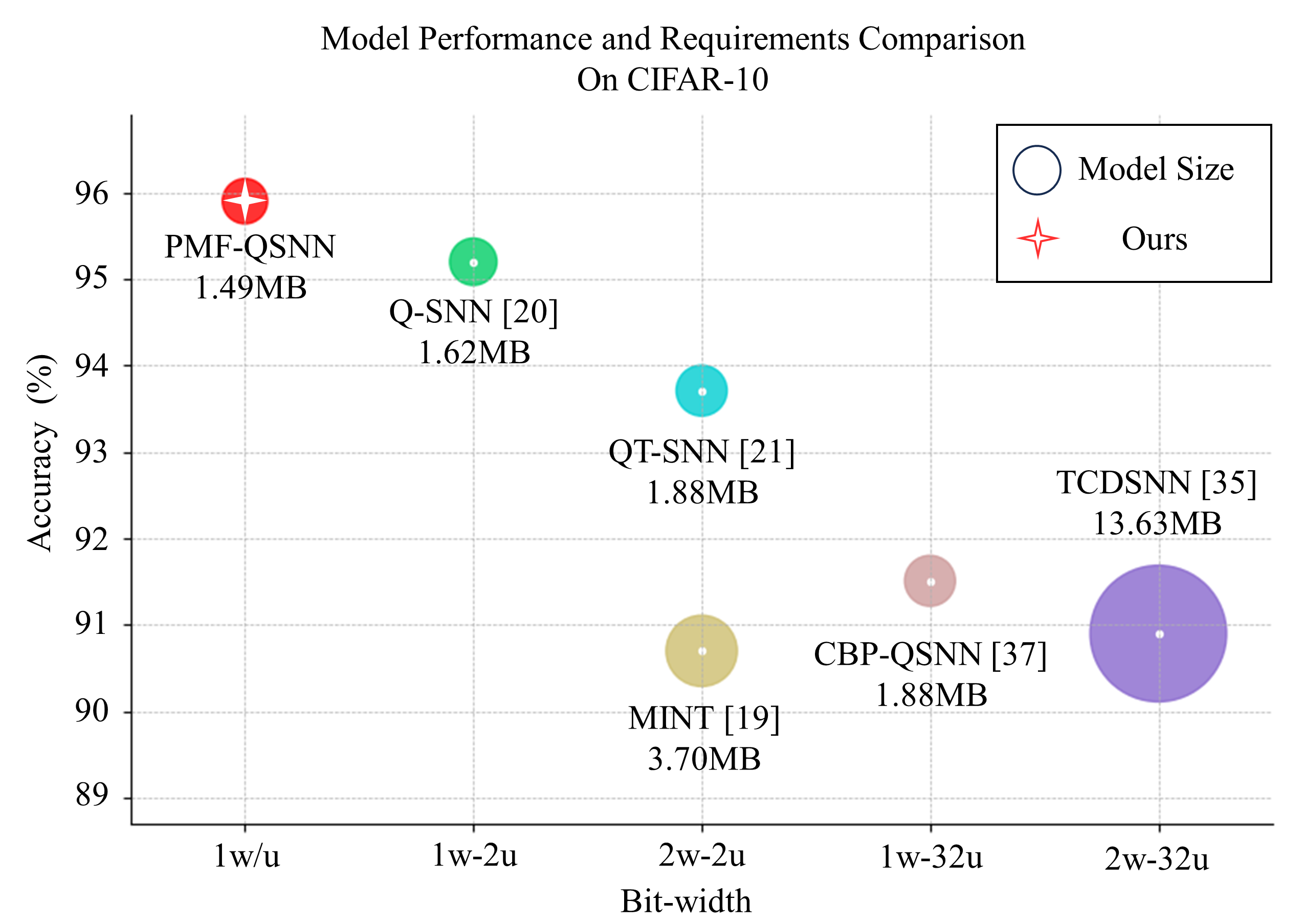}
    \centering
    \caption{A Comparative Analysis of Model Memory Requirements and Performance Across Various Methods on the CIFAR-10 Dataset: Our method attains a recognition accuracy of 95.9\% while 1.48MB memory footprints. 
      }
    \label{compare}
\end{figure}

Additionally, we compare the memory efficiency advantages of our model. As illustrated in Fig.~\ref{compare}, we showcase the memory requirements and performance of various quantization methods on the CIFAR10 dataset. It is evident that our model maintains competitive results with a lower memory footprint.

\subsection{Ablation Study}

To further validate the spatio-temporal interaction capabilities and training efficiency of the PMF-QSNN, we conducted ablation studies on the CIFAR10DVS dataset. Specifically, we utilize the VGGSNN architecture as the full-precision SNN baseline and verify the performance of MFP-QSNN, including accuracy, memory footprint, and training speed.

\begin{table}[ht]
\caption{Comparison of models' memory footprint and accuracy.}
\label{table2}
\centering
\begin{tabular}{ccccc}
\toprule
Methods               & Timesteps & Accuracy (\%) & Memory (MB)          & Speed (ms) \\ \midrule
Baseline              &  4     &  79.6   & 325.933 &     3652            \\ \midrule
\multirow{3}{*}{Ours} & 4  &   79.1 & 9.600                     &   1.5\(\times\)             \\ 
                      & 8     &  79.8        &              9.606        &  3.1\(\times\)              \\
                      & 10    &  81.1        &            9.611          & 7.2\(\times\)               \\ \bottomrule
\end{tabular}
\end{table}

As shown in Table.~\ref{table2}, we demonstrate the memory footprint of our model. Compared with a full-precision baseline, PMF-QSNN achieves a \(33\times\) optimization, requiring only 9.6MB for inference. Additionally, we achieve a 7.2\(\times\) increase in training speed when timesteps are 10 across 100 training epochs.

\section{Conclusion}

In this paper, we propose MFP-QSNN to address the significant performance degradation observed in QSNNs. Firstly, the memory-free quantization method does not require the storage of membrane potentials for historical information exchange, thereby significantly enhancing QSNN performance with reduced storage requirements. Additionally, parallel training and asynchronous inference processes further accelerate training speeds while ensuring asynchronous inference capability in SNNs. Extensive experiments demonstrate that our MFP-QSNN holds promise for improving efficient neuromorphic computing in resource-constrained environments.

\section{Acknowledgments}
This work was supported in part by the National Natural Science Foundation of China under grant U20B2063, 62220106008, and 62106038, the Sichuan Science and Technology Program under Grant 2024NSFTD0034 and 2023YFG0259.

\bibliographystyle{IEEEbib} 
\bibliography{my}

\end{document}